\documentclass[conference]{IEEEtran}
\IEEEoverridecommandlockouts

\usepackage{cite}
\usepackage{amsmath,amssymb,amsfonts}
\usepackage{algorithmic}
\usepackage{graphicx}
\usepackage{subfigure}
\usepackage{textcomp}
\usepackage{xcolor}
\usepackage{booktabs} 
\usepackage{array} 
\usepackage{algorithm} 
\usepackage{listings} 
 \usepackage{color, soul} 

\definecolor{codegreen}{rgb}{0,0.6,0}
\definecolor{codegray}{rgb}{0.5,0.5,0.5}
\definecolor{codepurple}{rgb}{0.58,0,0.82}
\definecolor{backcolour}{rgb}{0.95,0.95,0.92}

\lstdefinestyle{mystyle}{
	backgroundcolor=\color{backcolour},
	commentstyle=\color{codegreen},
	keywordstyle=\color{magenta},
	numberstyle=\tiny\color{codegray},
	stringstyle=\color{codepurple},
	basicstyle=\ttfamily\footnotesize,
	breakatwhitespace=false,
	breaklines=true,
	captionpos=b,
	keepspaces=true,
	numbers=left,
	numbersep=5pt,
	showspaces=false,
	showstringspaces=false,
	showtabs=false,
	tabsize=2
}
\def\BibTeX{{\rm B\kern-.05em{\sc i\kern-.025em b}\kern-.08em
		T\kern-.1667em\lower.7ex\hbox{E}\kern-.125emX}}

\begin{document}
	\title{PANAMA: A Network-Aware MARL Framework for Multi-Agent Path Finding in Digital Twin Ecosystems}
	\author{\IEEEauthorblockN{Arman~Dogru, R.~Irem~Bor-Yaliniz,~\IEEEmembership{Senior Member,~IEEE,} and Nimal~Gamini~Senarath}
		\IEEEauthorblockA{\textit{Huawei Canada Advanced Research Center} \\
			Ottawa, Canada \\
}
	}
	
	\maketitle
	
	\begin{abstract}
	Digital Twins (DTs) are transforming industries through advanced data processing and analysis, positioning the world of DTs, \textit{Digital World}, as a cornerstone of next-generation technologies including embodied AI. As robotics and automated systems scale, efficient data-sharing frameworks and robust algorithms become critical. We explore the pivotal role of data handling in next-gen networks, focusing on dynamics between application and network providers (AP/NP) in DT ecosystems. We introduce PANAMA, a novel algorithm with Priority Asymmetry for Network Aware Multi-agent Reinforcement Learning (MARL) based multi-agent path finding (MAPF). By adopting a Centralized Training with Decentralized Execution (CTDE) framework and asynchronous actor-learner architectures, PANAMA accelerates training while enabling autonomous task execution by embodied AI. Our approach demonstrates superior pathfinding performance in accuracy, speed, and scalability compared to existing benchmarks. Through simulations, we highlight optimized data-sharing strategies for scalable, automated systems, ensuring resilience in complex, real-world environments. PANAMA bridges the gap between network-aware decision-making and robust multi-agent coordination, advancing the synergy between DTs, wireless networks, and AI-driven automation.
	\end{abstract}
	
\section{Introduction}
Embodied artificial intelligence (Embodied AI) serves as a key enabler for achieving high levels of automation, as it enables physical entities, such as robots, to perceive and dynamically interact with their environment~\cite{embodied_AI}. 6G and beyond networks can play a key role for embodied AI to support communications, sensing and computation offloading~\cite{network_for_AI}. Another enabler can be the Digital Twins, which can provide substantial validation, testing and prediction capabilites for high-level automation in complex environments~\cite{dt_net, dt_ai}. At the intersection of both methods, lies data. In fact, as the 6G standardization unfolds, data framework of 3GPP, such as data handling capabilities and even the boundaries of 3GPP data are under discussion. Although agentic AI, personal connectivity assistance and several robotics-related use cases are already accepted in 3GPP 22.280 v031, data exposure issues encircling these applications remain ambiguous, making role of 6G questionable in enabling automation. One one hand, assume that network data is exposed to the application provider (AP) for the AP to derive necessary insights. On the other, assume that the AP data is exposed to the network provider (NP) for NP to derive related information. Neither one of these scenarios is favorable from both practical and regulatory aspects. In this paper, we argue that DTs can play a key role in creating buffer zones for data exposure and, by proposing a novel multi-agent reinforcement learning (MARL) based multi-agent path finding (MAPF) algorithm (PANAMA), we show how both NP and AP can benefit from a balanced data exposure, which increases communication efficiency and reliability simultaneously.

\begin{figure*}[htbp]
	\centerline{\includegraphics[width=.8\textwidth]{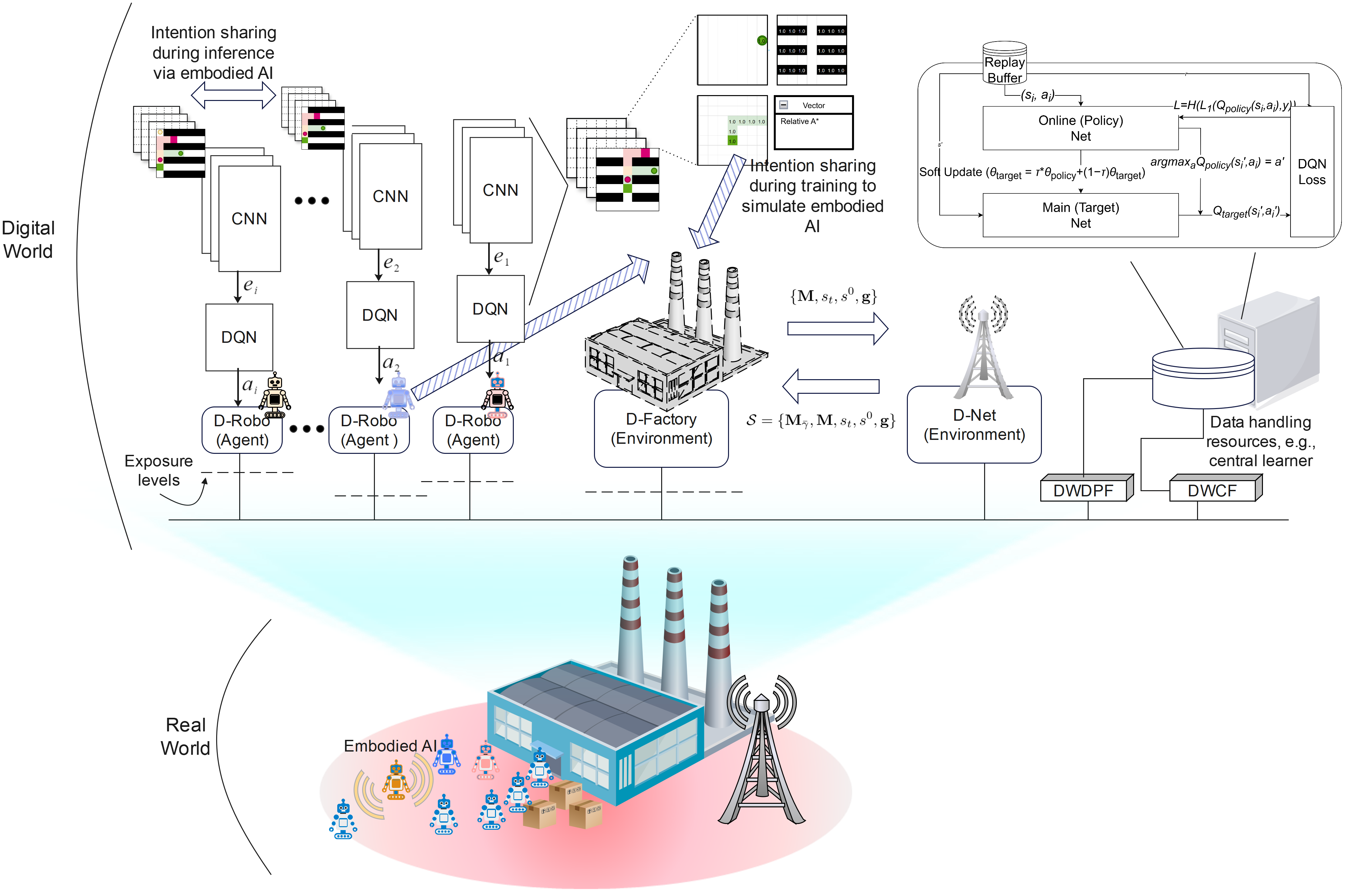}}
	\caption{The asynchronous actor-learner architecture. Multiple `Actor` processes collect experience in parallel and send it to the `Central Learner`. The learner optimizes the shared policy and broadcasts updated weights back to the actors.}
	\label{fig:arch_diagram}
\end{figure*}

First, we propose a DT ecosystem where DTs belonging to various real-world entities, such as humans, robots, and networks. DTs can collect data, provide information, control actuators, and interact with other DTs and real-world entities. In Fig.~\ref{fig:arch_diagram} we show a subset of this ecosystem with DTs of robots (D-Robot), a factory (D-Factory) and a network (D-Net). These DTs collaborate to train D-Robots and then use inference to test the performance of operation continuously. Note that networks provide integrated sensing and communications (ISAC) services to maintain DTs and enable analysis for the operation. The presented architecture can be used in several frameworks. For example, the DTs in this framework can be act as agents in an agentic AI operation in a fully automated warehouse scenario. Proposed Digital World Control Function (DWCF) and Digital World Data Processing Function (DWDPF) provide control and data plan functionalities, respectively, to enable dynamic creation and operation of DT networks (DTN). The benefits of this ecosystem become evident when scaled to solve complex and critical scenarios.

The cooperative MAPF problem is a fundamental challenge in robotics and AI with wide-ranging applications, from warehouse automation and traffic control to video games and drone swarm coordination~\cite{embodied_AI}. We formally model this complex, interactive decision-making scenario as a Decentralized Partially Observable Markov Decision Process (Dec-POMDP). 
In order to improve capability to scale up and comply with the proposed network architecture, we employ a Centralized Training with Decentralized Execution (CTDE) paradigm~\cite{embodied_AI}. Training of a central, mutual policy comes with challenges inherent in MARL, where the policy fails to co-operate with copies of itself. Our novel asymetric priority based observations force the observed state information to be asymetrical and facilitates cooperation with a centrally trained policy. Considering realistic communication constraints by utilizing D-Net, we finally obtain the PANAMA, which not only demonstrates network awareness but also has superior performance in terms of training efficiency and scalability compared to the state-of-the-art benchmarks.
\section{System Model}
\label{sec:Sys}
The first component of  the system model is the DT ecosystem and D-Net modeling. The second one is the real-world environment and how it is represented in the PANAMA framework. 
\subsection{Network Architecture}\label{sec:wireless_model}
6G networks can enable automation by improving data handling and connectivity capabilities, e.g., by supporting Model Context Protocol and improving anonymous communication capabilities. For example, DWCF can provide a network slice for the specific DTN, while DWDPF provide data processing capabilities. Fig.~\ref{fig:arch_diagram} shows how D-Robots share location and planned movement with D-Factory, and D-Factory provides FoV. In addition, D-Net receives D-Robot paths, factory map and other relevant information from D-Factory to provide corresponding network analysis. The data exposure is managed by DWCF and DWDPF, as well as internal DT functions, while satisfying privacy and security constraints. 

Ideally, D-Net is the digital twin of the wireless network, which represents not only the equipment but also channel and other complex entities. In this study, the D-Net comprises a physical layer model determining the signal quality across the map. We model a multi-sector cellular deployment where agents correspond to users. The received signal power in dBm, $P_{\text{rx,dB}}(i, s)$, for an agent $i$ from a base station sector $s$ is given by:
\begin{equation}
	P_{\text{rx,dB}}(i, s) = P_{\text{tx,dB}} + G_{\text{ant}}(i, s) - \text{PL}(i, s),
\end{equation}
where $P_{\text{tx,dB}}$ is the sector's transmit power in dBm, $G_{\text{ant}}(i, s)$ is the directional antenna gain in dBi from sector $s$ towards agent $i$, and $\text{PL}(i, s)$ is the total path loss in dB, based on 3GPP TR 36.837 for Urban Micro (UMi) environments, which includes a shadowing component. 

An agent $i$ is served by the sector providing the strongest signal: $s_i^* = \arg\max_s P_{\text{rx,dB}}(i, s)$. The received power in linear scale from any sector $s$ is $P_{\text{rx,lin}}(i, s) = 10^{P_{\text{rx,dB}}(i, s) / 10}$. The Signal-to-Interference-plus-Noise Ratio (SINR) for agent $i$ at time $t$, denoted as $\gamma_i(t)$, is then:
\begin{equation}
	\gamma_i(t) = \frac{P_{\text{rx,lin}}(i, s_i^*(t))}{\sum_{s' \neq s_i^*(t)} P_{\text{rx,lin}}(i, s') + N_0},
	\label{eq:sinr_linear}
\end{equation}
where the denominator sums the interference from all non-serving sectors and adds noise power, $N_0$. An agent's potential data rate, $R_i(t)$, is determined from its SINR using a standard Modulation and Coding Scheme (MCS) lookup table: $R_i(t) = \text{MCSLookup}(\gamma_i(t))$~\cite{mcs}. The SINR map, normalized to a $[0, 1]$ range as $\bar{\gamma}_i(t)$, is provided as the fourth channel in each agent's spatial observation.

\subsection{MARL Framework and Environment}
To allow for a fair comparison, our environment is designed to be similar to those used in state-of-the-art benchmarks~\cite{okumura_priority_2022, damani_primal2_2021}. We model the problem as a Decentralized Partially Observable Markov Decision Process (Dec-POMDP) for $N$ agents, indexed by $i \in \{1, \dots, N\}$.

\textbf{State Space ($\mathcal{S}$)}: The global state $s_t \in \mathcal{S}$ at timestep $t$ is a complete snapshot of the environment. It includes the positions and priorities of all agents, $s_t = \{(x_i(t), y_i(t), p_i(t))\}_{i=1}^N$, where $(x_i, y_i)$ are the grid coordinates and $p_i$ is the dynamic priority of agent $i$ (Sec.~\ref{sec:priority}). The problem definition also includes static elements: A binary map $\mathbf{M} = [\mathbf{m}_1, \mathbf{m}_2, \mathbf{m}_k, ..., \mathbf{m}_W]$ where $\mathbf{m}_k = [\mathbf{m}_{k,0},\mathbf{m}_{k,1},\mathbf{m}_{k,l}, ..., \mathbf{m}_{k,H}]^\mathbb{T}$ and $\mathbf{m}_{k,l} \in \{0, 1\}$ indicating impassable (1) cells, a normalized network map $\mathbf{M}_{\bar\gamma}$ where $( \mathbf{M}_{\bar\gamma} )_{k,l} \in [0,1]$ indicating normalized $\gamma_i(t)$ values of map tiles, and agent-specific start positions $\mathbf{s}^0 = [(x^0_0, y^0_0), (x^0_1, y^0_1), ..., (x^0_i, y^0_i)]$ and goals $\mathbf{s}^g = [(x^g_0, y^g_0), (x^g_1, y^g_1), ..., (x^g_i, y^g_i)]$

%

\textbf{Action Space ($\mathcal{A}$)}: Each agent $i$ has a discrete action space $\mathcal{A}_i = \{\text{up, down, left, right, stay}\}$. The joint action at timestep $t$ is $\mathbf{a}_t = (a_1(t), \dots, a_N(t))$, where $a_i(t) \in \mathcal{A}_i$. The environment enforces valid transitions; if multiple agents attempt to move to the same cell, the conflict resolution mechanism forces lower-priority agents to execute the `stay` action.

\textbf{Observation Space ($\mathcal{O}$)}: Execution is decentralized, so each agent $i$ receives a local observation $o_i(t) \in \mathcal{O}_i$. This observation is a tuple $o_i(t) = (\mathbf{O}_i(t), \mathbf{v}_i(t))$, designed for a dual-stream neural network.
\begin{itemize}
	\item A \textit{spatial tensor} $\mathbf{O}_i(t) \in \mathbb{R}^{4 \times V \times V}$, where $V$ is the side length of the agent's square field-of-view (FoV). The four channels are:
	\begin{enumerate}
		\item \textit{Static Obstacles}: A binary map of walls and other static impediments within the agent's FoV, derived from $\mathbf{M}$.
		\item \textit{Other Agents}: A binary map indicating the positions of other agents within the FoV.
		\item \textit{Communicated Intentions}: A map of planned future path segments from nearby \textit{higher-priority} agents.
		\item \textit{Network Awareness}: A map of the normalized SINR, $\bar{\gamma}_i(t)$, for traversable cells within the FoV.
	\end{enumerate}
	\item A \textit{feature vector} $\mathbf{v}_i(t)$, which includes non-spatial, strategic information:
	\begin{equation}
		\mathbf{v}_i(t) = (\mathbf{d}_i^g, f_i^g, \mathbf{d}_{i,1}^{A*}, \dots, \mathbf{d}_{i,k}^{A*}).
	\end{equation}
	This vector contains: the agent's relative coordinates to its goal $\mathbf{d}_i^g = (x_i^g - x_i(t), y_i^g - y_i(t))$; a boolean flag $f_i^g$ indicating if it has reached its goal; and the relative coordinates $\mathbf{d}_{i,k}^{A*}$ to the next $k$ waypoints on its pre-computed A* path.
\end{itemize}
The experience gathered by each agent $i$ at each step $t$ is stored as a transition tuple $(o_i(t), a_i(t), r_i(t+1), o_i(t+1), d_i(t+1))$. Here, $r_i(t+1)$ is the reward received after executing action $a_i(t)$, $o_i(t+1)$ is the resulting observation, and $d_i(t+1)$ indicates if the episode terminated. For brevity in subsequent sections, we denote a generic transition from agent $i$ as $(o_i, a_i, r_i, o'_i, d_i)$.
	\section{The PANAMA Method}
	\label{sec:Method}
The asynchronous actor-learner architecture of CTDE comprises one central learner process and multiple parallel actor processes, communicating via shared multiprocessing queues. 
The centralized entity aggregates transition from all agents into a shared Prioritized Experience Replay (PER) buffer. Using this global data-set, it trains a single, shared policy network. Hence,  centralized learner can correlate actions and outcomes across the entire system, facilitating the learning of complex and generalized strategies that would be difficult to discover with independent learners.
	
The trained policy is distributed to concurrent workers running in parallel. Each worker manages an independent instance of the environment and a set of agents within it. During an episode, each agent uses its local copy of the policy to select an action based solely on its own partial observation. This makes the execution phase highly scalable and suitable for real-world deployment where agents cannot rely on a central controller.
	
	Data flows between these components via three main multiprocessing queues:
	\begin{itemize}
	\item Actors push collected experience tuples to this queue. The learner continuously pulls from it to fill its replay buffer.
	\item The learner periodically pushes the latest policy network weights and the current exploration rate ($\epsilon_t$) to a set of queues, one for each actor. Actors check for updates at the end of their episode to refresh their local policy.
	\item After each episode, actors push a dictionary of summary statistics (e.g., episode length, total reward) to this queue for logging and analysis in the training process.
	\end{itemize}
The agent's goal is to learn an optimal action-value function, $Q^*(o_t, a_t)$\cite{bellman} which estimates the maximum expected future return for taking action $a_t$ given local observation $o_t$ in a time step $t$. For weight initialization, our framework employs Kaiming (He) uniform initialization for all weights and sets all biases to zero. We utilize  \textbf{Double DQN (DDQN)} to mitigate maximization bias ~\cite{DDQN}  $\theta$ indicates the \textit{policy network}  and $\theta^-$ indicates a stable \textit{target network}. The target value ($y_i$) for a transition $(o_i, a_i, r_i, o'_{i})$ is calculated as:
	\begin{equation}
		y_i^{\text{DDQN}} = r_i + \gamma Q(o'_{i}, \underset{a'}{\text{argmax}} Q(o'_{i}, a'; \theta); \theta^-),
		\label{eq:ddqn}
	\end{equation}
	where $a'$ indicates the best action for the next observed state $o'$. This process involves:
	\begin{enumerate}
		\item \textbf{Action Selection}: Use the current policy network $\theta$ to find the best action $a'$ in the next observation $o'_{i}$.
		\item \textbf{Action Evaluation}: Use the stable target network $\theta^-$ to evaluate the Q-value of that chosen action.
		\item \textbf{Temporal Difference}: The final target is the immediate reward plus the discounted future value from the target network.
	\end{enumerate}
	
	Standard experience replay samples transitions uniformly, but not all experiences are equally valuable. Therefore, the \textbf{Prioritized Experience Replay (PER)} improves learning efficiency by sampling transitions more frequently based on the amount of TD error, $|\delta_i| = |y_i - Q(o_i, a_i; \theta)|$. The priority $p_i$ of an experience is calculated as $p_i = (|\delta_i| + \epsilon_{PER})^\alpha$, where $\alpha$ controls the degree of prioritization and $\epsilon_{PER}$ is a small constant to ensure non-zero transition priority. To correct for the bias introduced by non-uniform sampling, our framework employs Importance-Sampling (IS) , 
	$w_j = \left(\frac{1}{B} \cdot \frac{1}{P(j)}\right)^\beta,$
	where $P(j) = p_j / \sum_k p_k$ is the probability of sampling experience $j$, $B$ is the buffer size, and $\beta$ is a hyperparameter that anneals from an initial value (e.g., 0.4) towards 1.0 over the course of training. This annealing gradually increases the correction as the policy becomes more stable.
	
	To train the policy network, we sample a mini-batch of \(M\) transitions \(\{(o_i, a_i, r_i, o'_i, d_i)\}_{i=1}^M\) from the replay buffer, with associated importance‐sampling weights \(w_i\). For each sample \(i\), we form the Double DQN target:
	\begin{equation}
	y_i^{\mathrm{DDQN}}
	= r_i + \gamma\,(1 - d_i)\,
	Q\bigl(o'_i, \arg\max_{a'}Q(o'_i,a';\theta);\;\theta^{-}\bigr),
	\end{equation}
	where \(\theta^{-}\) are the target‐network parameters and $d_i$ is a boolean indicating if the episode terminated at that step.
	
	We then compute the per‐sample Huber~\cite{huber_robust_1964} loss between \(y_i^{\mathrm{DDQN}}\) and the current Q‐value \(Q(o_i,a_i;\theta)\). With a threshold \(\delta\) (typically 1.0). Then, total loss over the minibatch is the weighted average of the per-sample losses:
	\begin{equation}
		\mathcal{L}(\theta)
		= \frac{1}{M}\sum_{i=1}^{M} w_i\,\mathcal{L}_{\delta}\bigl(Q(o_i,a_i;\theta),\,y_i^{\mathrm{DDQN}}\bigr).
		\label{eq:ddqn_loss}
	\end{equation}
	This loss is backpropagated through the policy network, and its parameters \(\theta\) are updated using the AdamW optimizer (with weight decay). We apply gradient clipping (norm \(\le\) 1.0) to stabilize training. 
	

%
%
%
%
%
%
%
To maintain a stable learning target, the target network $\theta^-$ is updated slowly using a soft update via Polyak averaging after each learning step, $\theta^- \leftarrow \tau \theta + (1 - \tau) \theta^-$,	where $\tau$ is a small hyperparameter, set to 0.005 in our experiments. This ensures that the target values change slowly and smoothly, preventing oscillations and divergence in the learning process.

\textbf{Curriculum Learning} incrementally increases task difficulty to achieve a generalizable policy. The model trains on a sequence of stages with increasing complexity. This progression involves increasing the number of agents or moving to more challenging map layouts. Graduation to the next stage is automatically triggered when performance on the current stage, measured during periodic inference runs, surpasses a predefined success rate threshold, $\Gamma_c$, where $c$ indicates the curriculum stage. Upon graduating, the replay buffer is re-initialized to prevent data pollution from previous tasks. We periodically pause training and run evaluations using a greedy policy ($\epsilon \approx 0$). The aggregated success rate is compared against $\Gamma_c$ to decide if the agent is ready to advance~\cite{mnih_playing_nodate}.
\subsection{Environment and Behavioral Shaping}
\subsubsection{Problem Formulation}
\label{sec:problem_formulation}
At the beginning of each episode the environment's reset function intentionally creates deadlock scenarios by assigning agent pairs with randomized and swapped start and goal locations. This forces the policy to learn resolution strategies beyond simple path following.

\subsubsection{Asymmetrical Observation Space and Dynamic Priority System}
\label{sec:priority}
A key innovation is the \textbf{asymmetrical observation system}, which is crucial for breaking mirrored state observations that can lead to deadlocks in a shared policy execution. This system enables coordinated behavior. As described in Algorithm~\ref{alg:obs}, while an agent observes the locations of \textit{all} other agents in its field of view, it only sees the planned future path segments of \textit{higher-priority} agents. This asymmetry creates an implicit, dynamic hierarchy that encourages agents to yield to higher priority agents
To enable asymmetrical observations and conflict resolution, PANAMA uses a \textbf{dynamic priority} scheme,  re-calculated at every timestep. An agent's priority is its current A* distance to its goal (lower distance means higher priority), Note that once an agent reaches its goal it is demoted to the lowest priority, encouraging it to learn to move out of the way if it is blocking another active agent from reaching its own goal. This prevents ``goal camping" and is essential for achieving high success rates in congested scenarios.

\begin{algorithm}
	\caption{Asymmetric Observation Construction}
	\label{alg:obs}
	\begin{algorithmic}[1]
		\STATE \textbf{function} GetObservation(\texttt{agent})
		\STATE \texttt{o} $\leftarrow$ get obstacles in agent's FOV
		\STATE \texttt{a} $\leftarrow$ zeros\_like(\texttt{o})
		\STATE \texttt{c} $\leftarrow$ zeros\_like(\texttt{o})
		\FOR{\texttt{other\_a} in all agents}
		\IF{\texttt{other\_a} is in agent's FOV}
		\STATE Mark \texttt{other\_a.pos} in \texttt{a}
		\ENDIF
		\IF{\texttt{other\_a.p < agent.p}}
		\FOR{\texttt{path\_p} in \texttt{other\_a.comm\_p}}
		\IF{\texttt{path\_p} is in agent's FOV}
		\STATE Mark \texttt{path\_pos} in \texttt{c}
		\ENDIF
		\ENDFOR
		\ENDIF
		\ENDFOR
		\STATE \texttt{v} $\leftarrow$ [goal\_coords, is\_done, path\_coords...]
		\STATE \textbf{return} \{'spatial': [\texttt{o}, \texttt{a}, \texttt{c}], 'vector': \texttt{v}\}
	\end{algorithmic}
\end{algorithm}

\subsection{Reward Engineering}
\label{sec:reward_engineering}
The reward function is meticulously designed to guide agents toward cooperative and efficient behaviors. It is a sum of several components, detailed in Table~\ref{tab:rewards}. It includes a large sparse reward for reaching the goal, small penalties for time and collisions, and a dense potential-based reward shaping (PBRS) term for making progress. This term is calculated as $r_t^{\text{PBRS}} = \gamma\Phi(s_{t+1}) - \Phi(s_t)$, where the potential $\Phi(s_t) = -\beta \cdot d_{\text{A}^*}(s_t, g)$ is the scaled, negative A* distance from state $s_t$ to the goal $g$. A lower-priority agent is penalized with a \textit{path conflict penalty} for occupying a cell on a higher-priority agent's communicated path, encouraging cooperative yielding behavior. Finally, network-related rewards are included via a penalty proportional to poor signal quality, calculated as $r_{i,t}^{\text{net}} = (1 - \bar{\gamma}_i(t)) \cdot \lambda_{\text{penalty}}$, where $\bar{\gamma}_i(t)$ is the normalized SINR for agent $i$ (see Sec.~\ref{sec:wireless_model}) and $\lambda_{\text{penalty}}$ is a negative factor.

\begin{table}[htbp]
	\caption{Reward Function Components and Their Purpose}
	\begin{center}
		\begin{tabular}{l l >{\raggedright\arraybackslash}p{3.2cm}}
			\toprule
			\textbf{Reward Component} & \textbf{Value} & \textbf{Description \& Rationale} \\
			\midrule
			\texttt{reward\_goal} & +2.0 & Primary objective signal. \\
			\texttt{reward\_time\_step} & -0.01 & Penalty per step. \\
			\texttt{reward\_collision} & -1.0 & Penalty for any collision. \\
			\texttt{PBRS\_factor} & +0.23 & A* potential-based reward. \\
			\texttt{path\_conflict\_penalty} & -1.0 & Penalty for path conflict. \\
			\texttt{network\_penalty\_factor} & -0.4 & Penalty based on SINR.\\
			\bottomrule
		\end{tabular}
	\end{center}
	\label{tab:rewards}
\end{table}

\section{Experiments and Results}
\begin{figure*}[hbtp]  \centering  \subfigure[]{\includegraphics[width=0.24\textwidth]{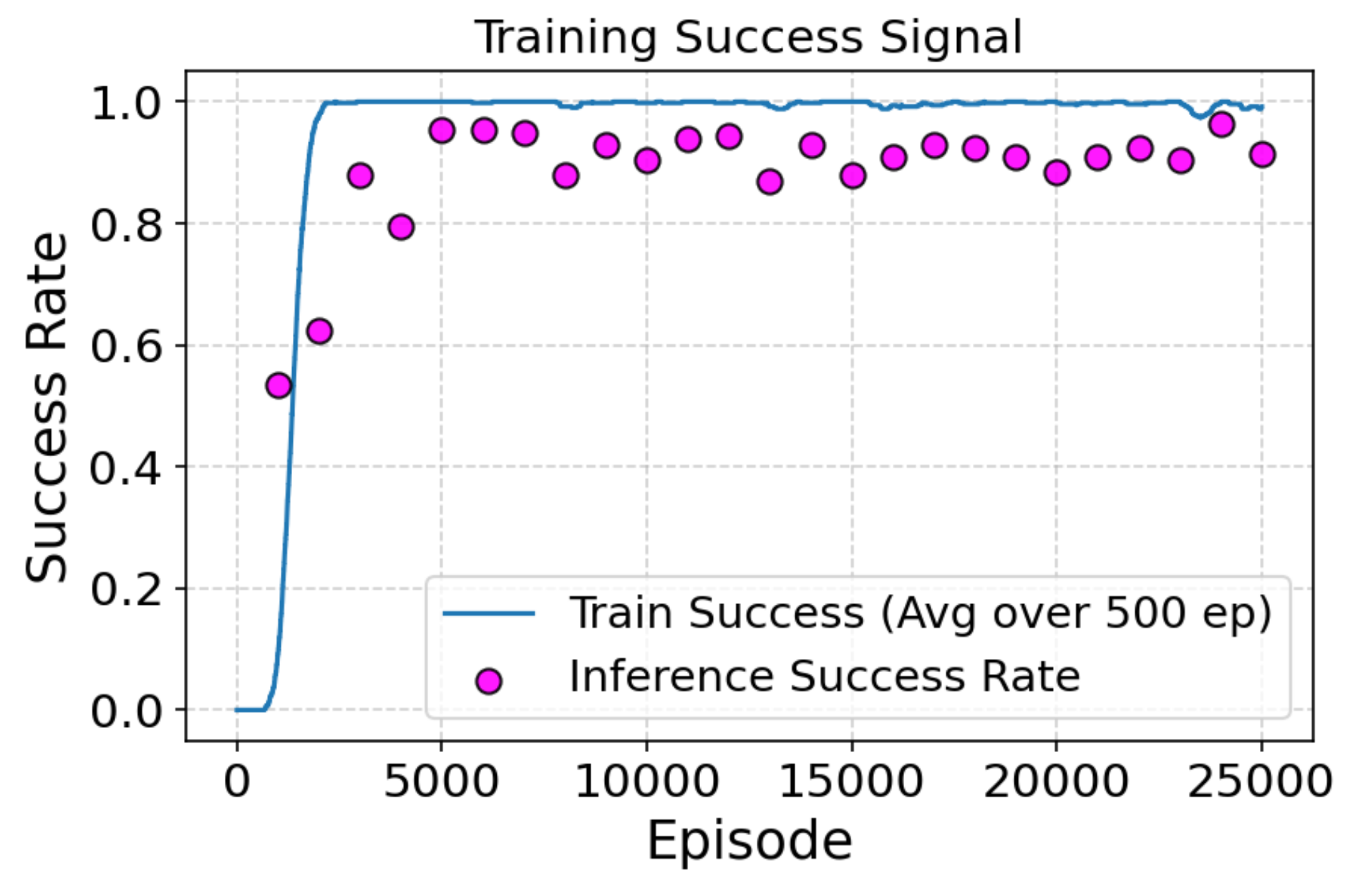}\label{d1}}  \subfigure[]{\includegraphics[width=0.24\textwidth]{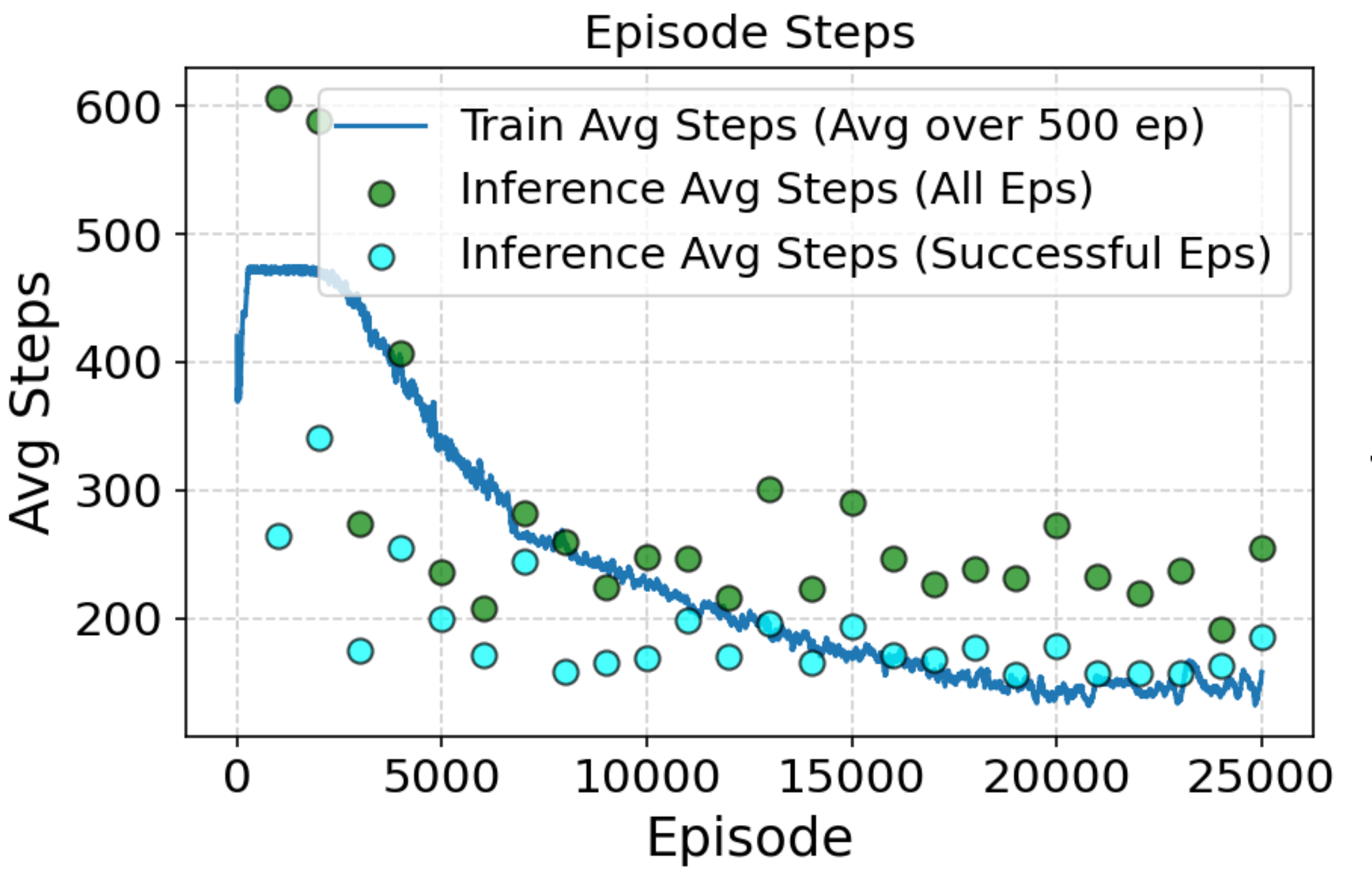}\label{d2}}   \subfigure[]{\includegraphics[width=0.21\textwidth]{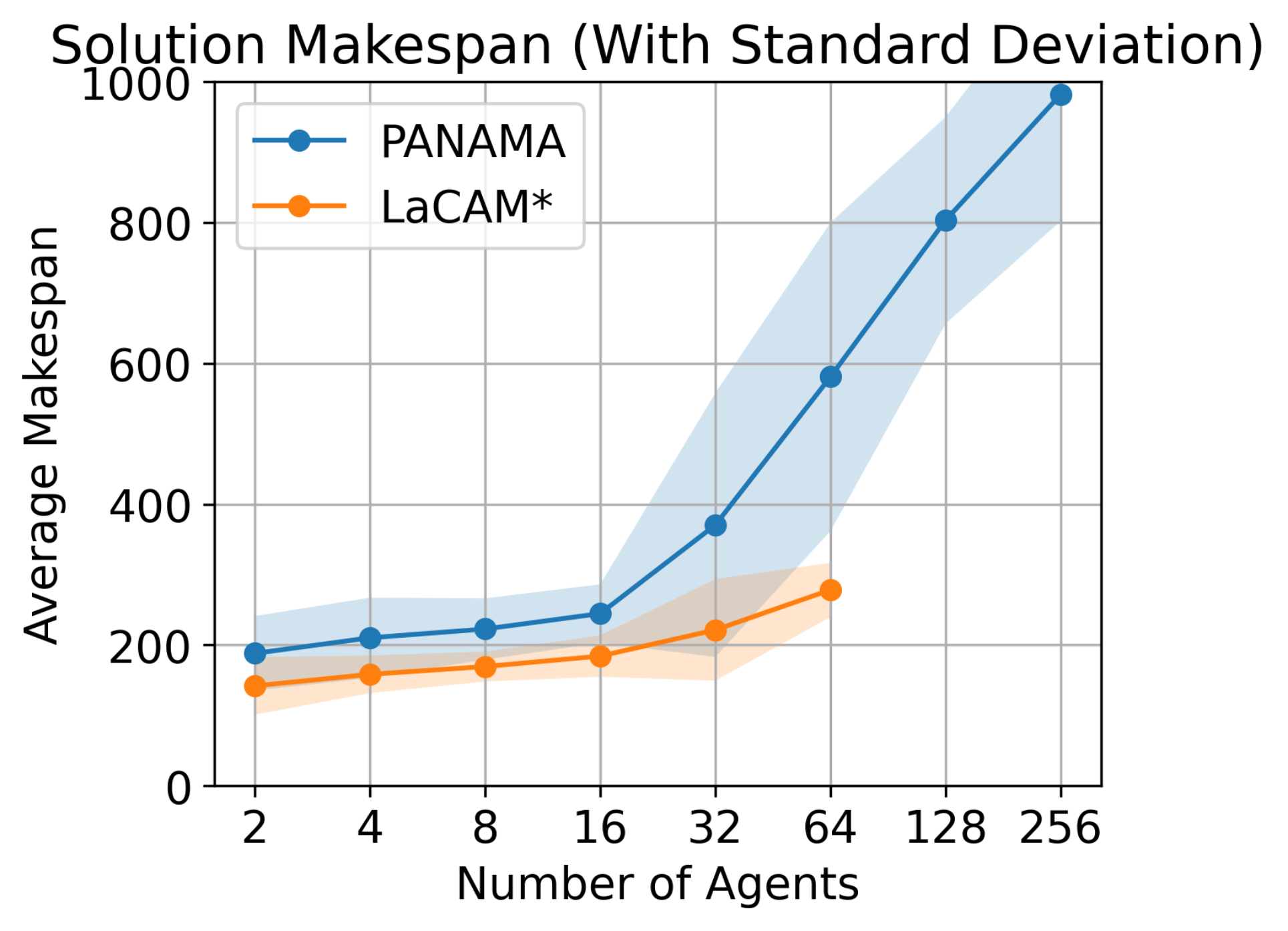}\label{c1}}  \subfigure[]{\includegraphics[width=0.18\textwidth]{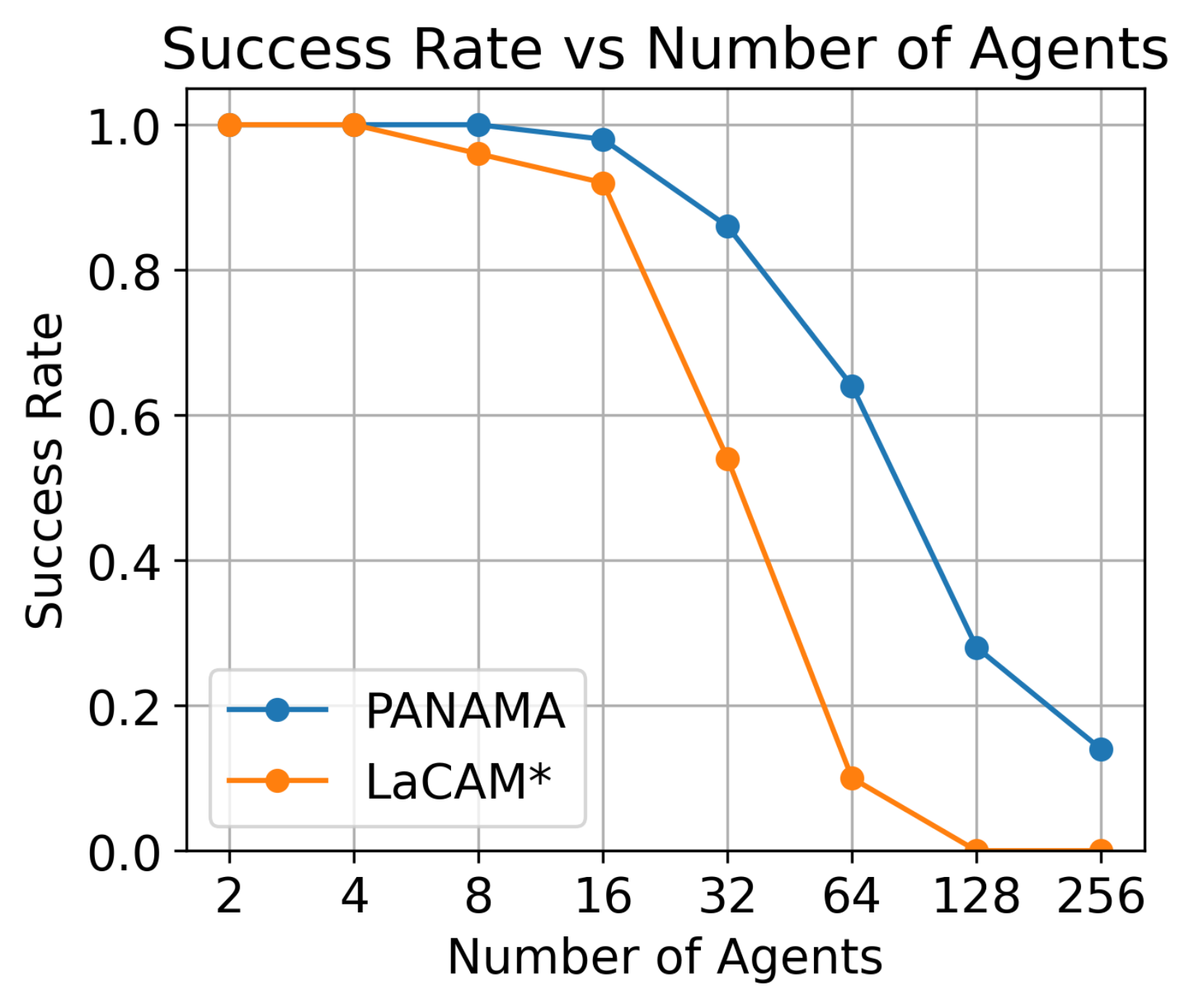}\label{c2}} \caption{Training and Inference metrics for Stage 5 (8 agents, Warehouse map). }  \label{fig:res_stage5}\end{figure*}

The parameters for D-Net, MARL, and curriculum schedule are detailed in Tables~\ref{tab:network_params},~\ref{tab:rl_params}, and~\ref{tab:curriculum}, respectively. Environmental parameters of D-Net and MARL are fine tuned to reflect the characteristics of warehouses and rooms. 
\label{sec:Experiments}
\begin{table}[htbp]
		\caption{D-Net Network Simulation Parameters}
		\label{tab:network_params}
		\begin{center}
			\begin{tabular}{l l >{\raggedright\arraybackslash}p{3cm}}
				\toprule
				\textbf{Value} & \textbf{Description} \\
				\midrule
				46.0 & BS transmit power (dBm). \\
				2.0 & Carrier frequency (GHz). \\
				10e6 & Channel bandwidth (10 MHz). \\
				17.0 & Max BS antenna gain (dBi). \\
				 8.0 & Antenna vertical downtilt. \\
				 65.0 & Horizontal 3dB beamwidth. \\
				10.0 & Vertical 3dB beamwidth. \\
				35.0 & BS antenna height (m). \\
				1.5 & Agent antenna height (m). \\
				10.0 & LoS shadowing std. dev. (dB). \\
				10.0 & NLoS shadowing std. dev. (dB). \\
				 37.0 & LoS decorrelation distance (m). \\
				50.0 & nLoS decorrelation distance (m). \\
				5.0 & BS noise figure (dB). \\
				7.0 & Agent noise figure (dB). \\
				\bottomrule
			\end{tabular}
		\end{center}
	\end{table}
	\begin{table}[htbp]
		\caption{MARL Parameters}
		\begin{center}
			\begin{tabular}{l l >{\raggedright\arraybackslash}p{3.8cm}}
				\toprule
				\textbf{Value} & \textbf{Description} \\
				\midrule
				0.99 & Discount factor. \\
				0.005 & Update rate for target network. \\
				1.0 & Initial exploration rate. \\
				0.1 & Final exploration rate. \\
				5,000,000 & Steps for epsilon to decay. \\
				1e-4 & AdamW learning rate. \\
				64 & Experiences per learning step. \\
				2,000,000 & Replay buffer size. \\
				0.6 & Prioritization exponent. \\
				0.4 & IS weight exponent. \\
				7 & Radius of agent's FOV (15x15). \\
				5 & A* steps in vector observation. \\
				5 & Future path steps communicated. \\
				2.0 & Multiplier on \texttt{fov\_r}, comm range. \\
				\bottomrule
			\end{tabular}
		\end{center}
		\label{tab:rl_params}
	\end{table}
%
	\begin{table}[ht]
		\caption{Curriculum Schedule}
		\begin{center}
			\begin{tabular}{llllllllll}
				\toprule
				\textbf{Stage} & \textbf{Name} & \textbf{Map Type} & \textbf{\# Agents} & \textbf{\# Steps}\\
				\midrule
				0 & Random-32x32-2 & Random & 2 & 200 \\
				1 & Random-32x32-4 & Random & 4 & 200\\
				2 & Room-32x32-4 & Room & 4 & 300 \\
				3 & Room-32x32-6 & Room & 6 & 300 \\
				4 & Room-32x32-8 & Room & 8 & 300 \\
				5 & Warehouse-161x63-8 & Warehouse & 8 & 1000\\
				\bottomrule
			\end{tabular}
		\end{center}
		\label{tab:curriculum}
	\end{table}
The framework was trained for 11 hours following the curriculum schedule and successfully graduated. Fig.~\ref{fig:res_stage5} shows that the training success rate steadily climbs towards 1.0. The inference success rate shows a consistent improvement, quickly surpassing the 90\% graduation threshold around episode 5,000. Initially, episodes finish around 450 steps. As the agents learn to cooperate in this new map, the average steps required to solve the problem decreases significantly, stabilizing around 100 steps. 

Furthermore, we conducted tests on MovingAI Benchmark Scenarios~\cite{stern2019mapf} with differing agent numbers across 50 test scenarios without network awareness for a fair comparison with LaCAM*~\cite{okumura_lacam_2023}  based on success rate and solution length (makespan) (Figure~\ref{fig:res_stage5}(c) and (d)). The learned policy demonstrates robust generalization and the success rate remains high while outperforming LaCAM* in scenarios with upwards of 256 agents. This highlights the scalability of the PANAMA.
	
\begin{figure*}[hbtp]  \centering  \subfigure[]{\includegraphics[width=.8\textwidth]{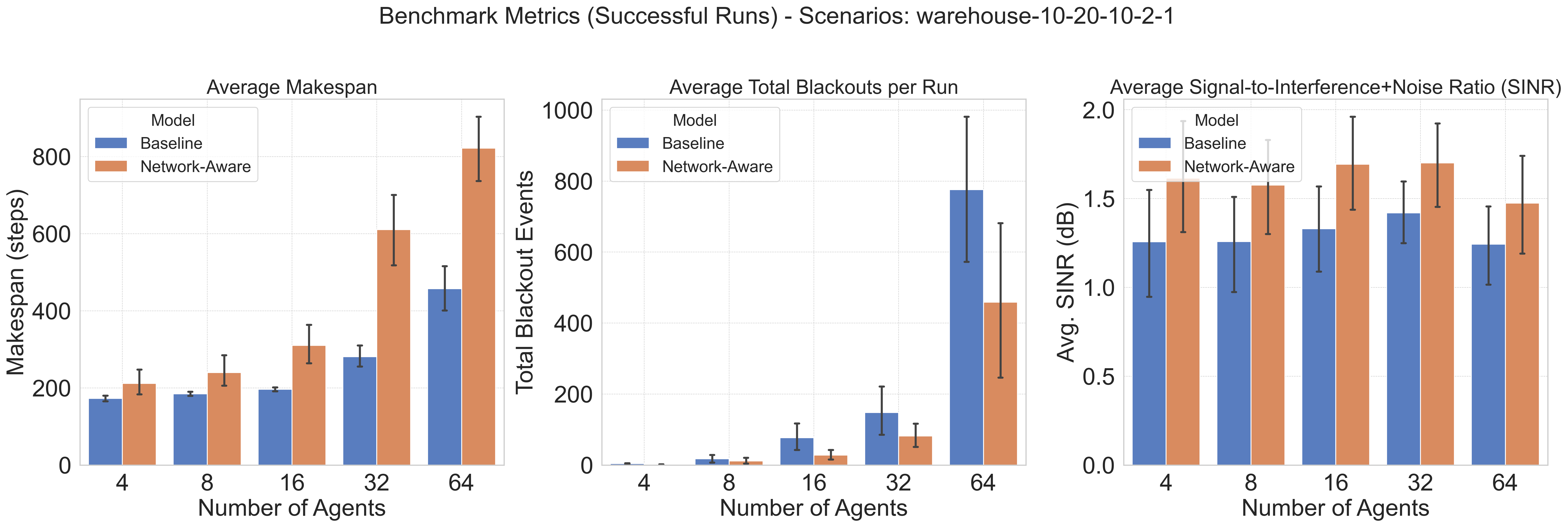}\label{d23}}  \subfigure[]{\includegraphics[width=.8\textwidth]{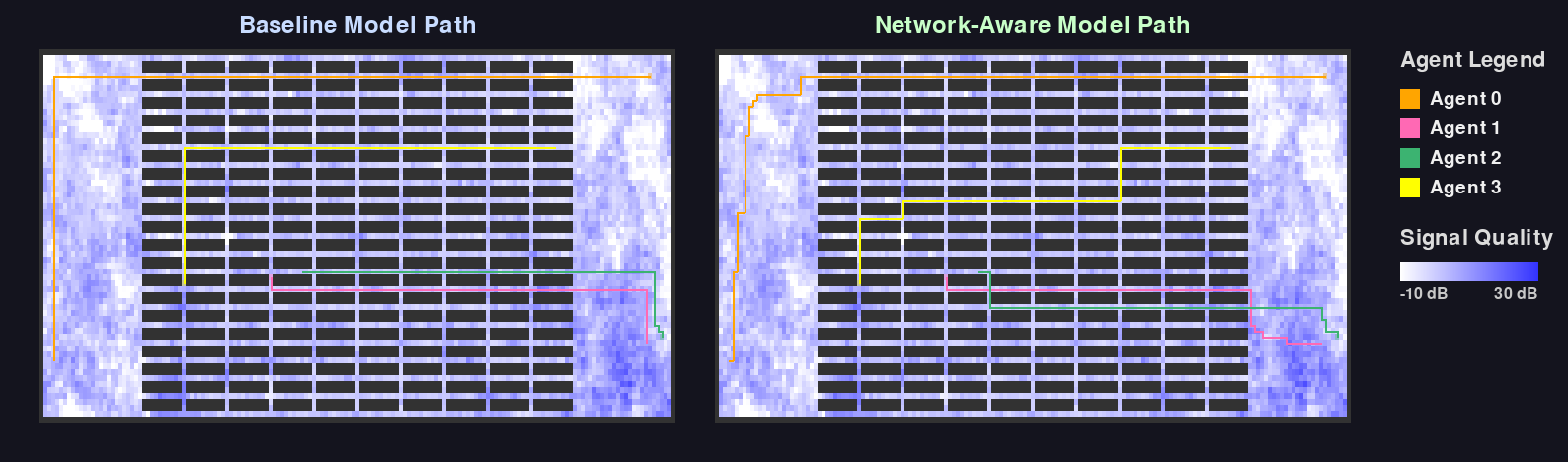}\label{d24}}  \caption{Monte Carlo experiments on "Warehouse-161x63" map by MovingAI. }  \label{fig:comparison}\end{figure*}

Finally,  we demonstrate path-planning and network usage of PANAMA with and without network awareness as shown in~Fig.\ref{fig:comparison}. Blackout events occur when the $\gamma_i(t)$ of agent $i$ is lower than the threshold $\kappa=-8.47$ to transmit any data (Sec.~\ref{sec:wireless_model}). Fig.~\ref{fig:comparison}(a) shows the trade-off between makespan and maintaining sufficient network performance. While the higher SINR and robustness against blackout is maintained by PANAMA even for the congested scenarios, makespan increases to find paths with good coverage. For example, Fig.~\ref{fig:comparison}(b) shows 4 agents who changes their path substantially due to network awareness.
\section{Conclusions and Future Work}
\label{sec:Conc}
PANAMA shows the trade-off between path selection and maintaining communication quality, along with insufficiency of benchmarks to scale up number of agents in challenging environments. We show the importance of network awareness for congested scenarios, where maintaining SINR and preventing blackouts become challenging. These results can provide insights for 6G and beyond networks to be desinged to support DT ecosystems allowing secure and privacy-preserving data exchange between APs and NPs. PANAMA can be further improved to support heterogenous agents and to be integrated in an agent DTN as planned in our future work.  
\section*{Acknowledgment}
Authors would like to thank Dr.~Jiayin Chen for their insightful discussions.

\bibliography{references.bib}
\bibliographystyle{IEEEtran}

\end{document}